\documentclass{article}
\usepackage{ijcai25}

\usepackage{times}
\usepackage{soul}
\usepackage{url}
\usepackage[hidelinks]{hyperref}
\usepackage[utf8]{inputenc}
\usepackage[small]{caption}
\usepackage{graphicx}
\usepackage{amsmath}
\usepackage{amsthm}
\usepackage{booktabs}
\usepackage[switch]{lineno}

\usepackage{subfigure}
\usepackage[ruled,vlined]{algorithm2e}
\usepackage{algpseudocode}
\usepackage{pifont}
\usepackage{subcaption}
\usepackage{amsfonts}
\usepackage{multirow}
\usepackage{bm}
\usepackage[table]{xcolor}

\newcommand{\eg}{\textit{e}.\textit{g}.}

\title{Physical Adversarial Camouflage through Gradient Calibration and Regularization}

\author{
Jiawei Liang$^{1,2}$\and
Siyuan Liang$^{3*}$\and
Jianjie Huang$^{1}$\and
Chenxi Si$^1$\and
Ming Zhang$^4$\and
Xiaochun Cao$^{1,2}$\thanks{Corresponding author.}\\
\affiliations
$^1$School of Cyber Science and Technology, Sun Yat-sen University Shenzhen Campus, China\\
$^2$Peng Cheng Laboratory, Shenzhen, China\and
$^3$Nanyang Technological University, Singapore\\
$^4$National Key Laboratory of Science and Technology on Information System Security, Beijing, China\\
\emails
\{liangjw57, huangjj67, sichx\}@mail2.sysu.edu.cn, siyuan.liang@ntu.edu.sg,\\ zm\_stiss@163.com, caoxiaochun@mail.sysu.edu.cn
}

\begin{document}

\maketitle

\begin{abstract}

The advancement of deep object detectors has greatly affected safety-critical fields like autonomous driving. However, physical adversarial camouflage poses a significant security risk by altering object textures to deceive detectors. Existing techniques struggle with variable physical environments, facing two main challenges: 1) inconsistent sampling point densities across distances hinder the gradient optimization from ensuring local continuity, and 2) updating texture gradients from multiple angles causes conflicts, reducing optimization stability and attack effectiveness. To address these issues, we propose a novel adversarial camouflage framework based on gradient optimization. First, we introduce a gradient calibration strategy, which ensures consistent gradient updates across distances by propagating gradients from sparsely to unsampled texture points. Additionally, we develop a gradient decorrelation method, which prioritizes and orthogonalizes gradients based on loss values, enhancing stability and effectiveness in multi-angle optimization by eliminating redundant or conflicting updates. Extensive experimental results on various detection models, angles and distances show that our method significantly exceeds the state of the art, with an average increase in attack success rate (ASR) of 13. 46\% across distances and 11.03\% across angles. Furthermore, empirical evaluation in real-world scenarios highlights the need for more robust system design.

\end{abstract}
\section{Introduction}\label{sec: introduction}

Object detection~\cite{zou2023object,ren2016faster,liang2024object,liang2023exploring} is essential for applications like autonomous driving, where robustness and reliability~\cite{liang2022large,liang2024poisoned,liang2025vl} are crucial due to its integration into safety-critical systems. Adversarial attacks~\cite{liang2021generate,wei2018transferable} pose a serious threat by manipulating object textures to deceive detection algorithms. This can lead to serious potential real-world consequences~\cite{muxue2023adversarial,kong2024patch} such as accidents.

Physical adversarial camouflage~\cite{wang2022fca,liu2023x,kong2024environmental} presents significant challenges compared to its digital counterparts~\cite{liang2020efficient,liang2022parallel}, as it requires modifying real-world objects in a manner that ensures the attack remains effective when transitioning from the digital domain to uncontrolled physical environments. This process involves optimizing an object's surface texture to deceive detection systems across varying angles and distances, aligning with the inherent variability in how cameras perceive objects under diverse perspectives. However, achieving robust adversarial camouflage necessitates addressing several critical challenges. One key difficulty stems from the localized nature of texture manipulation. Specifically, from any given viewpoint, only the visible portions of the object's texture can be optimized. This viewpoint-specific focus often neglects the broader context of the object's surface, which may lead to suboptimal performance when interactions across different perspectives are considered holistically. Furthermore, inconsistencies in the object's appearance, caused by variations in viewing distance, angle, or environmental conditions, further compound the complexity of this task.

\begin{figure}[t]
    \centering
    \includegraphics[width=\columnwidth]{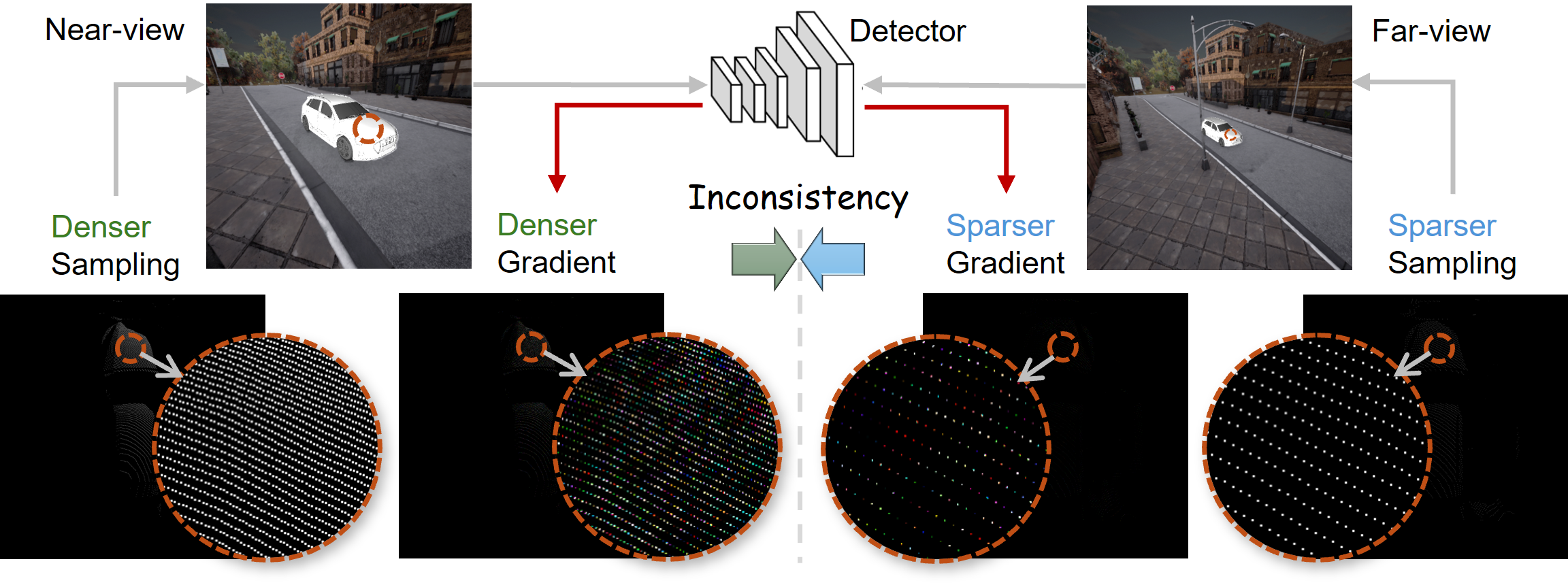}
    \caption{Illustration of gradient inconsistencies due to variations in sampling density across different distances.}
    \vspace{-0.3cm}
    \label{fig: distance issue}
\end{figure}

\begin{figure}[t]
    \centering
    \includegraphics[width=0.95\columnwidth]{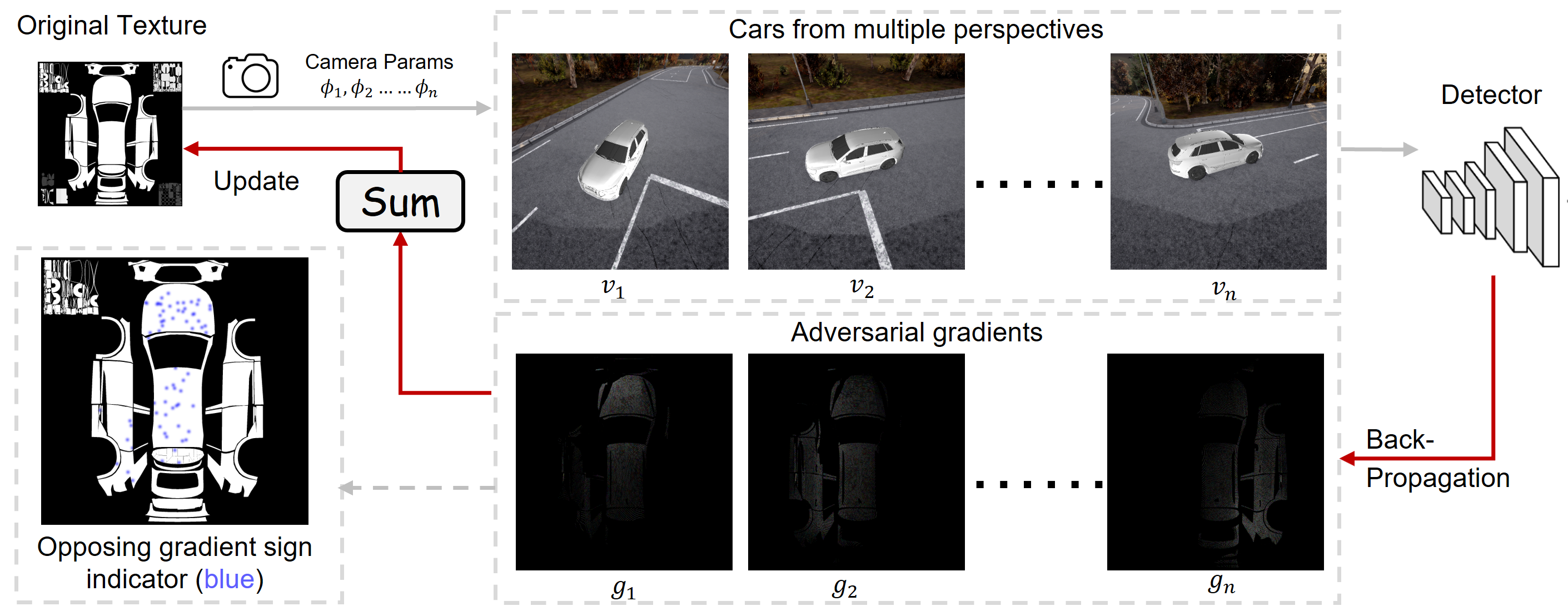}
    \caption{Illustration of gradient conflicts from different perspectives}
    \vspace{-0.3cm}
    \label{fig: angle issue}
\end{figure}

Existing methods typically utilize differentiable renderers to optimize adversarial textures by simulating object appearances across diverse viewpoints and environmental conditions. However, we identify two critical issues within this optimization pipeline that adversely affect attack performance. The first issue arises from distance-dependent sampling density: as the distance between the camera and the object changes, the number of pixels used to render the object varies, as illustrated in Figure~\ref{fig: distance issue}. This variation leads to inconsistent gradient sparsity during backpropagation. This inconsistency, in turn, results in uneven texture modifications across different distances. The second issue stems from potential redundancy or conflict among gradient directions within a minibatch, as illustrated in Figure~\ref{fig: angle issue}. Gradients derived from similar viewpoints tend to generate redundant updates, while gradients from distinct viewpoints may contradict and cancel one another, thus impeding effective texture optimization.

To address these identified challenges, we propose a novel adversarial camouflage framework built upon two core strategies: Nearest Gradient Calibration (NGC) and Loss-Prioritized Gradient Decorrelation (LPGD). First, to mitigate the issue of inconsistent gradient sparsity, NGC propagates gradients from sampled points to neighboring unsampled points on the same surface, ensuring local continuity and consistent sparsity in texture updates across varying distances. Second, to resolve the gradient redundancy and conflicts arising from diverse viewpoints within a minibatch, LPGD prioritizes gradients according to their associated adversarial loss and decorrelates them through orthogonalization. This orthogonalization process ensures that each prioritized gradient provides complementary and non-conflicting information, thereby overcoming challenges related to redundancy and cancellation.

We conducted experiments in realistic simulated environments, demonstrating that our adversarial camouflage effectively evades object detection across diverse viewpoints, distances, weather conditions, and multiple detectors. It outperforms state-of-the-art methods by 13.46\% in attack effectiveness across distances and 11.03\% across angles. Real-world experiments further confirm its practicality and robustness.

\textbf{Our contributions are as follows:}
\begin{itemize}
    \item We identify two challenges in optimizing adversarial camouflage, including inconsistent gradient sparsity due to distance-dependent sampling density, and redundant or conflicting gradients across viewpoints.
    \item We propose Nearest Gradient Calibration (NGC) to propagate gradients from sampled to unsampled texture points, ensuring local continuity in gradient updates across varying distances.
    \item We introduce Loss-Prioritized Gradient Decorrelation (LPGD) to prioritize and decorrelate gradients via orthogonalization, resolving redundancy and conflicts across viewpoints.
    \item Extensive experiments demonstrate that our approach achieves superior attack performance, enhanced robustness to environmental variations, and improved transferability across diverse object detectors compared to state-of-the-art methods.
\end{itemize}

\section{Related Work}\label{sec: related work}

\subsection{Object Detection}
Object detection plays a pivotal role in computer vision, evolving through various strategies. Anchor-based methods include two-stage detectors~\cite{girshick2015fast,he2017mask,ren2015faster}, which use region proposals for classification and bounding box regression, and single-stage models~\cite{liu2016ssd,redmon2018yolov3}, which frame detection as a regression problem to achieve real-time performance. Conversely, anchor-free methods~\cite{duan2019centernet} eliminate predefined anchors by directly predicting object centers and dimensions. Recent developments like DETR~\cite{zhu2020deformable} introduce an end-to-end detection approach utilizing transformer architectures. Nevertheless, these methods continue to be susceptible to adversarial attacks.

\subsection{Adversarial Attacks}
Adversarial attacks involve the deliberate alteration of images to mislead deep neural networks~\cite{szegedy2013intriguing}, resulting in incorrect predictions. These attacks are divided into digital and physical based on the domain of perturbation. Digital attacks directly apply crafted perturbations to digital images, with techniques such as FGSM~\cite{goodfellow2014explaining} and PGD~\cite{mkadry2017towards} demonstrating notable success, especially in classification. These strategies have been adapted for object detection but lose effectiveness in real-world scenarios. In contrast, physical attacks~\cite{wei2024physical,zhu2023tpatch} introduce perturbations in the real world, encompassing adversarial patches and camouflage. Patch-based attacks apply perturbations to flat surfaces, effective only from certain viewpoints, while adversarial camouflage covers entire object surfaces, allowing for multi-angle attacks.

\begin{figure*}[thbp]
    \centering
    \includegraphics[width=\textwidth]{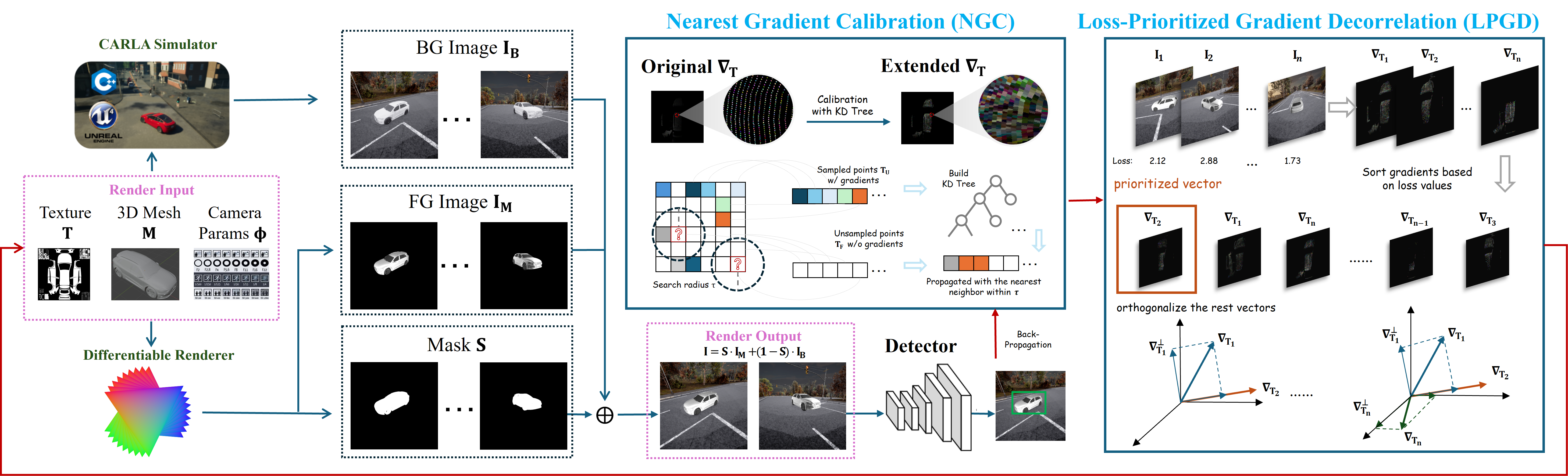}
    \caption{Overview of our proposed method. To optimize the texture, we first render the target vehicle image using the texture with a renderer and integrate it into a realistic background generated by the CARLA Simulator. During optimization, we employ NGC to calibrate gradients in sparsely sampled regions and use LPGD to resolve gradient conflicts through loss-prioritized orthogonalization on a minibatch. The refined gradients are then summed and utilized to update the texture.}
    \label{fig: framework}
    \vspace{-0.4cm}
\end{figure*}

\subsection{Adversarial Camouflage}
Current adversarial camouflage methods often utilize differentiable 3D renderers to project 3D objects as rendered images from multiple angles, optimizing 3D textures end-to-end via backpropagation. For instance, Wang et al.~\cite{wang2022fca} introduced Full-coverage Camouflage Attack (FCA), optimizing a vehicle's UV textures through a neural renderer for enhanced multi-viewpoint camouflage. Similarly, the Dual Attention Suppression (DAS) attack by Wang et al.~\cite{wang2021dual} reduces visibility to both models and human observers. Another approach involves optimizing 2D texture patterns projected onto vehicle surfaces. Suryanto et al.~\cite{Suryanto_2022_CVPR} proposed the Differentiable Transformer Attack (DTA), employing a neural renderer to simulate realistic effects like shadows. Additionally, ACTIVE~\cite{Suryanto_2023_ICCV} improves camouflage aesthetics with enhanced texture mapping and background color integration. Furthermore, RAUCA~\cite{zhou2024rauca} utilizes advanced rendering to account for environmental factors such as diverse weather conditions. However, existing techniques may not fully resolve challenges arising from sampling inconsistency and conflicts in gradient updates, which can affect overall attack effectiveness.

\section{Method}\label{sec: method}

\subsection{Problem Definitions}

Adversarial camouflage~\cite{wei2024physical} aims to optimize the texture of a target object to achieve objectives such as evading detection or inducing misclassification. A 3D object is represented by the tuple $(\mathbf{M}, \mathbf{T})$, where $\mathbf{M}$ denotes the mesh structure and $\mathbf{T}$ represents its texture. The object's image, rendered under various camera parameters $\varphi \in \mathbf{\Phi}$ (\eg, different camera angles and distances), is produced using a differentiable renderer, $\mathcal{R}$. The rendering process first computes the UV coordinates as a function of the camera parameter $\varphi$, defined by:
\begin{equation}
    \mathbf{UV}(\varphi) = \mathcal{R}(\mathbf{M}, \varphi).
\label{equ: rendering}
\end{equation}
The final image of the target object is obtained by sampling color values from the texture $\mathbf{T}$ at the computed UV coordinates:
\begin{equation}
    \mathbf{I}_{\mathbf{M}}(\mathbf{T}, \varphi) = \mathcal{F}_{S}(\mathbf{UV}(\varphi), \mathbf{T}),
\label{equ: sampling}
\end{equation}
where $\mathcal{F}_{S}$ denotes the sampling function. Since the renderer $\mathcal{R}$ does not incorporate background information, a common approach for creating physically realistic images is to overlay the rendered object onto authentic backgrounds. In this study, we utilize background images generated by the CARLA simulator~\cite{dosovitskiy2017carla}. The composite image is created by replacing the background in the rendered image using a segmentation mask $\mathbf{S}$:
\begin{equation}
    \mathbf{I}(\mathbf{T}, \varphi) = (\mathbf{S} \cdot \mathbf{I}_{\mathbf{M}}(\mathbf{T}, \varphi)) + (1 - \mathbf{S}) \cdot \mathbf{I}_{\mathbf{B}},
\label{equ:combining_fg_bg}    
\end{equation}
where $\mathbf{I}_{\mathbf{B}}$ represents the background image.

To conduct a physical evasion attack, the texture is optimized to mislead the target model, causing detection failure or misclassification. The adversarial texture is derived by solving:
\begin{equation}
    \mathbf{T}_{\text{adv}} = \arg\min_{\mathbf{T}} \mathcal{L}\big(\mathbb{F}_{\theta}(\mathbf{I}(\mathbf{T}, \varphi)),\ y\big),
\label{equ: general_attack}
\end{equation}
where $\mathbb{F}_{\theta}$ is the object detector with parameters $\theta$, $y$ is the ground-truth label covering classification and localization, and $\mathcal{L}$ is the suppression loss function. A lower loss implies a reduced probability of predicting the target label correctly. The objective is to hinder the model's ability to accurately predict the target label $y$.

\textbf{Existing Obstacles:} Achieving effective adversarial camouflage in multi-view settings requires accounting for variations in distance and viewing angles. Current methods~\cite{wang2022fca,zhou2024rauca} optimize textures by rendering images under diverse conditions using differentiable renderers~\cite{kato2018neural}. However, they fail to address two significant challenges: \ding{182} Varying sampling densities at different distances. This variation causes the gradients used for texture updates to exhibit inconsistent levels of sparsity across distances. \ding{183} Conflicting texture update directions across different viewing angles. Naively merging these updates undermines the effectiveness of the attack.

\subsection{Nearest Gradient Calibration}

As illustrated in Equation~\ref{equ: sampling}, the rendered image samples colors from the texture \(\mathbf{T}\) using UV coordinates. The pixel count representing the target object varies with rendering distance: closer objects occupy more pixels, resulting in denser texture sampling, whereas distant objects are represented by fewer pixels, leading to sparser sampling. Typically, only sampled points are updated during texture optimization. Due to varying rendering distances producing different sets of sampled points, inconsistencies in texture updates can emerge across different distances. This issue arises from emphasizing pixel-level optimization while neglecting the continuity required to preserve the integrity of local surface regions.

To address this problem, we propose a simple yet effective method called \textbf{Nearest Gradient Calibration} (NGC). The core idea of NGC is to extend the gradient of each sampled point to its neighboring unsampled points. This propagation ensures that gradient updates over the texture remain locally continuous, thereby preserving the consistency of texture updates across varying distances.

We define the trainable region of the texture \( \mathbf{T} \) as \( \mathbf{T}' \), which is constrained by a mask \( \mathbf{K} \) that restricts updates to specific areas, such as the visible outer surface of a vehicle. Within \( \mathbf{T}' \), the sampled subset of points is denoted as \( \mathbf{T}_{\text{U}}(\varphi) \), determined by the camera parameters \( \varphi \). For simplicity, we refer to \( \mathbf{T}_{\text{U}}(\varphi) \) as \( \mathbf{T}_{\text{U}} \) in the following sections. The remaining unsampled points are defined as \( \mathbf{T}_{\text{F}} = \mathbf{T}' \setminus \mathbf{T}_{\text{U}} \). Formally, this relationship is expressed as:
\begin{equation}
    \mathbf{T}' = \mathbf{T} \odot \mathbf{K} = \mathbf{T}_{\text{U}} \cup \mathbf{T}_{\text{F}}
\label{equ: trainable texture}
\end{equation}

After each iteration of backpropagation, only the gradients of the sampled points, \(\nabla_{\mathbf{T}_\text{U}}\), are computed.
\begin{equation}
    \nabla_{\mathbf{T}_\text{U}} = \nabla_{\mathbf{T}_\text{U}} \mathcal{L} \big( \mathbb{F}_{\theta} (\mathbf{I}(\mathbf{T}, \varphi)),\ y \big )
\end{equation}
Ideally, adjacent points on the same surface should be updated collectively to maintain continuity. However, the renderer's sampling mechanism may result in sparsely distributed gradients, leaving intermediate points without gradients. To address this, we propagate the sparse gradients to their neighboring unsampled points using a nearest neighbor search, implemented via the \textit{KD Tree} algorithm~\cite{zhou2008real}. For each unsampled point \( p \in \mathbf{T}_\text{F} \), we calculate the Euclidean distance to all points in \(\mathbf{T}_\text{U}\) to identify the nearest neighbor \( q \). The gradient of the nearest sampled point \( q \in \mathbf{T}_\text{U} \) is assigned to \( p \) provided that \( \| p - q \|_2 \leq \tau \), where \( \tau \) is the search radius. This ensures that gradients are assigned only to unsampled points within a specified local range, maintaining spatial locality. If no sampled point within the threshold is found, the gradient for \( p \) remains zero. Formally, the gradient assignment is defined as:
\begin{equation}
\begin{aligned}
    \forall p \in \mathbf{T}_\text{F}, \quad
    \nabla_{\mathbf{T}_\text{F}}^{p} &= 
    \begin{cases} 
        \nabla_{\mathbf{T}_\text{U}}^{q}, & \text{if } \| p - q \|_2 \leq \tau, \\
        0, & \text{otherwise},
    \end{cases} \\
    \text{where } q &= \arg\min_{q' \in \mathbf{T}_\text{U}} \| p - q' \|_2.
\end{aligned}
\end{equation}
After assigning gradients to \(\mathbf{T}_\text{F}\), we obtain the final extended gradient $\nabla_{\mathbf{T}'}$. By applying NGC, gradients are smoothly propagated across the texture surface, ensuring local continuity, as illustrated in Figure~\ref{fig: framework}. This approach effectively addresses the inconsistencies caused by varying sampling densities, resulting in more uniform and coherent texture updates across different rendering distances. The detailed procedure for this method is summarized in Algorithm~\ref{alg:Optimized_NGC_2}.

\begin{algorithm}[t]
\SetAlgoLined
\SetKwInOut{Input}{Input}
\SetKwInOut{Output}{Output}

\Input{Texture \(\mathbf{T}\), Mask \(\mathbf{K}\), camera parameter set \(\Phi\), Search radius \(\tau\)}
\Output{Extended gradients \(\nabla_{\mathbf{T}'}\)}
\BlankLine

Define \(\mathbf{T}' = \mathbf{T} \odot \mathbf{K}\)\;
Initialize KD-Tree \(\mathcal{K}\) with \(\mathbf{T}_{\text{U}}\)\;
\For{each optimization step}{
    Sample camera parameter \(\varphi \in \Phi\)\;
    Identify sampled points \(\mathbf{T}_{\text{U}}(\varphi)\)\;
    
    \(\mathbf{T}_{\text{F}} \gets \mathbf{T}' \setminus \mathbf{T}_{\text{U}}\)\;
    
    Compute gradients \(\nabla_{\mathbf{T}_{\text{U}}}\)\;
    
    Query nearest neighbors for \( p \in \mathbf{T}_{\text{F}} \) using \(\mathcal{K}\)\;
    
    \For{each \( p \in \mathbf{T}_{\text{F}} \)}{
        Retrieve nearest \(q\) and distance \(d\)\;
        
        \eIf{$d \leq \tau$}{
            Assign \(\nabla_{\mathbf{T}_\text{F}}(p) \gets \nabla_{\mathbf{T}_{\text{U}}}(q)\)\;
        }{
            Assign zero gradient: \(\nabla_{\mathbf{T}_\text{F}}(p) \gets 0\)\;
        }
    }
}
\Return \(\nabla_{\mathbf{T}'}\)\;
\caption{\fontsize{9.7}{9.7}\selectfont Nearest Gradient Calibration (NGC)}
\label{alg:Optimized_NGC_2}
\end{algorithm}

\subsection{Loss-Prioritized Gradient Decorrelation}

In multi-view physical adversarial optimization, overlapping texture regions observed from varying azimuth and elevation angles present substantial challenges. These challenges primarily manifest as gradient redundancy and conflicts. Gradient redundancy arises when gradients from different viewpoints exhibit similar update directions, leading to inflated gradient norms that destabilize optimization. Conversely, gradient conflicts occur when gradients act in opposing directions, resulting in partial or complete cancellation of updates and diminishing their effectiveness. If unaddressed, these issues can potentially reduce efficiency of adversarial optimization.

To address these problems, we propose the \textbf{Loss-Prioritized Gradient Decorrelation} (LPGD) method, which integrates gradient orthogonalization with prioritization based on loss values. This method ensures that the optimization focuses on the most challenging viewpoints and resolves redundancy and conflicts among gradients to produce stable and efficient updates.

In the LPGD method, gradients from \(k\) viewpoints, corresponding to the trainable part of the texture \(\mathbf{T}'\) in Equation~\ref{equ: trainable texture}, are first sorted according to their loss values, \(\mathcal{L}(\varphi_i)\), such that \(\mathcal{L}(\varphi_1) \geq \mathcal{L}(\varphi_2) \geq \dots \geq \mathcal{L}(\varphi_k).\)
This prioritization ensures that gradients from more challenging viewpoints are processed first. Specifically, the gradients \(\{\nabla_{\mathbf{T}'}(\varphi_1), \nabla_{\mathbf{T}'}(\varphi_2), \dots, \nabla_{\mathbf{T}'}(\varphi_k)\}\) are ordered in this manner before applying gradient orthogonalization.

Once gradients are prioritized, we proceed with gradient orthogonalization to resolve redundancy and conflicts. This step is crucial to ensure that each gradient contributes unique, non-redundant information to the optimization process. We apply a Schmidt orthogonalization~\cite{leon2013gram} procedure, which projects each gradient onto the orthogonal complement of the subspace spanned by the previously processed gradients. Formally, given the \(i\)-th gradient \(\nabla_{\mathbf{T}'}(\varphi_i)\), its orthogonalized counterpart \(\nabla_{\mathbf{T}'}^{\perp}(\varphi_i)\) is computed as:

\begin{equation}
\begin{aligned}
\nabla_{\mathbf{T}'}^{\perp}(\varphi_i) &= \nabla_{\mathbf{T}'}(\varphi_i) - \sum_{j=1}^{i-1} \alpha_{ij} \nabla_{\mathbf{T}'}^{\perp}(\varphi_j), \\
\text{where } \alpha_{ij} &= \frac{\nabla_{\mathbf{T}'}(\varphi_i) \cdot \nabla_{\mathbf{T}'}^{\perp}(\varphi_j)}{\|\nabla_{\mathbf{T}'}^{\perp}(\varphi_j)\|^2}.
\end{aligned}
\end{equation}
This recursive projection removes any components of \(\nabla_{\mathbf{T}'}(\varphi_i)\) that are redundant or conflicting with respect to previously orthogonalized gradients, ensuring that all gradients are mutually decorrelated. As a result, the gradients provide complementary information about the texture updates, rather than conflicting or overlapping contributions.

Once all gradients are orthogonalized, the final update direction is determined by averaging the orthogonalized gradients. The final update direction is computed as:
\begin{equation}
    \nabla_{\mathbf{T}'}^{*} = \frac{1}{k} \sum_{i=1}^k \nabla_{\mathbf{T}'}^{\perp}(\varphi_i),
\end{equation}
where \(\nabla_{\mathbf{T}'}^{\perp}(\varphi_i)\) are the orthogonalized gradients. By combining orthogonalization with averaging, the method achieves stable updates that respect the contributions of all viewpoints without being dominated by redundant or conflicting gradients. The detailed procedure for this method is summarized in Algorithm~\ref{alg:loss_prioritized_optimization}.

Notably, our proposed NGC and LPGD methods are compatible and can be combined in a straightforward manner. NGC propagates the gradient of points \(q \in \mathbf{T}_{\text{U}}\), i.e., \(\nabla_{\mathbf{T}_{\text{U}}}\), to points \(p \in \mathbf{T}_{\text{F}}\), i.e., \(\nabla_{\mathbf{T}_{\text{F}}}\), to obtain the extended gradient \(\nabla_{\mathbf{T}'}\) at each iteration. Subsequently, LPGD decorrelates these extended gradients \(\nabla_{\mathbf{T}'}\) within a minibatch via loss-prioritized orthogonalization to obtain the final gradients \(\nabla_{\mathbf{T}'}^{*}\).

\begin{algorithm}[t]
\SetAlgoLined
\SetKwInOut{Input}{Input}
\SetKwInOut{Output}{Output}

\Input{Viewpoints \( \{\varphi_i\}_{i=1}^k \), corresponding gradients \( \{\nabla_{\mathbf{T}'}(\varphi_i)\}_{i=1}^k \), and losses \( \{\mathcal{L}(\varphi_i)\}_{i=1}^k \)}
\Output{Optimized gradient \( \nabla_{\mathbf{T}'}^{*} \)}
\BlankLine

Sort \( \nabla_{\mathbf{T}'}(\varphi_i) \) in descending order of \( \mathcal{L}(\varphi_i) \)\;

Initialize \( \nabla_{\mathbf{T}'}^{\perp}(\varphi_1) \gets \nabla_{\mathbf{T}'}(\varphi_1) \)\;

\For{each \( i = 2 \) to \( k \)}{
    
    \For{each \( j = 1 \) to \( i-1 \)}{
        Compute \(
        \alpha_{ij} = \frac{\nabla_{\mathbf{T}'}(\varphi_i) \cdot \nabla_{\mathbf{T}'}^{\perp}(\varphi_j)}{\|\nabla_{\mathbf{T}'}^{\perp}(\varphi_j)\|^2}
        \)
    }

    Update \(\nabla_{\mathbf{T}'}^{\perp}(\varphi_i) \gets \nabla_{\mathbf{T}'}(\varphi_i) - \sum_{j=1}^{i-1} \alpha_{ij} \nabla_{\mathbf{T}'}^{\perp}(\varphi_j)
    \)
}

Compute \( \nabla_{\mathbf{T}'}^{*} \gets \frac{1}{k} \sum_{i=1}^{k} \nabla_{\mathbf{T}'}^{\perp}(\varphi_i) \)

\Return \( \nabla_{\mathbf{T}'}^{*} \)\;
\caption{Loss-Prioritized Gradient Decorrelation (LPGD)}
\label{alg:loss_prioritized_optimization}
\end{algorithm}

\section{Experiment}\label{sec: experiment}

 
\subsection{Settings}

\textbf{Implementation Details}:We utilize the CARLA simulator~\cite{dosovitskiy2017carla} to generate datasets. In line with previous studies~\cite{zhou2024rauca}, we capture simulated images to construct a comprehensive training set. Our training set comprises 20,000 images captured from diverse angles and distances to enhance texture generation. We focus on the Audi E-Tron model, as explored in prior research~\cite{wang2021dual,wang2022fca,Suryanto_2022_CVPR,Suryanto_2023_ICCV}. For evaluation, adversarial camouflage is applied to the vehicle within CARLA, with images captured at elevation angles of $\{0^\circ, 5^\circ, 10^\circ, 15^\circ, 20^\circ, 30^\circ, 45^\circ, 60^\circ\}$, along with 2-degree azimuth increments for a thorough $360^\circ$ sweep. Extended evaluations encompass distances of $\{5, 7.5, 10, 12.5, 15\}$ meters and five distinct weather conditions: noon, sunset, night, foggy, and rainy. In real-world tests, the camouflage patterns are printed on 1:24 scale Audi E-Tron models, with images captured from various angles and distances for further analysis. Texture optimization is conducted using the Adam optimizer with a learning rate of 0.1, employing ModernGL~\cite{Dombi2020} for differentiable rendering and segmentation mask $\mathbf{S}$ generation. We optimize the textures over three epochs, with all experiments performed on a single NVIDIA A100 80GB GPU.

\begin{figure}[t]
\begin{center}
\centerline{\includegraphics[width=\columnwidth]{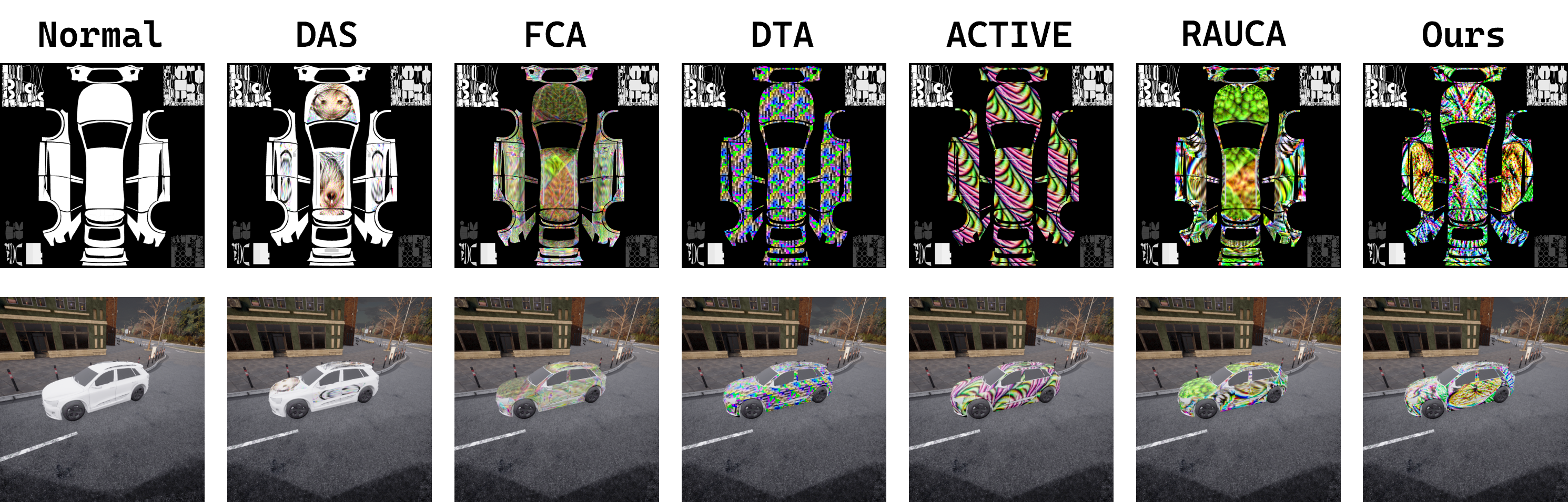}}
\caption{Optimized textures for each baseline method and the corresponding rendered samples.}
\vspace{-0.7cm}
\label{fig: texture}
\end{center}
\end{figure}

\textbf{Comparison Baselines}: Our framework is evaluated against several state-of-the-art adversarial camouflage techniques, including DAS~\cite{wang2021dual}, FCA~\cite{wang2022fca}, DTA~\cite{Suryanto_2022_CVPR}, ACTIVE~\cite{Suryanto_2023_ICCV}, and RAUCA~\cite{zhou2024rauca}. 
For our evaluations, as illustrated in Figure~\ref{fig: texture}, we transform both our optimized textures and those of the baselines into standardized UV textures, which are constrained by a mask $\mathbf{K}$ applied to the optimizable regions.

\textbf{Evaluation Metrics}\label{evaluation-metrics}: Following previous studies~\cite{zhou2024rauca}, we assess the attack effectiveness of adversarial camouflage using the AP@0.5 benchmark \cite{everingham2015pascal}, which is a standard measure capturing both recall and precision at a detection IOU threshold of 0.5.

\textbf{Target Detection Models}\label{target-models}: Consistent with prior studies~\cite{zhou2024rauca}, we utilize YOLOv3~\cite{redmon2018yolov3} as the white-box target detection model for generating adversarial camouflage. To evaluate the effectiveness of the optimized camouflage, we test it against a range of popular object detection models, treating them as black-box models, except for YOLOv3. This evaluation encompasses models such as the one-stage detector YOLOX~\cite{ge2021yolox}, two-stage detectors Faster R-CNN (FrRCN)~\cite{ren2016faster} and Mask R-CNN (MkRCN)~\cite{he2017mask}, as well as transformer-based detectors DETR~\cite{carion2020end} and PVT~\cite{wang2021pyramid}. Each model is pretrained on the COCO dataset and implemented using the MMDetection framework~\cite{chen2019mmdetection}.

\subsection{Evaluation in Physically-Based Simulation Settings}

In this section, we compare our method to state-of-the-art (SOTA) adversarial camouflage approaches.
Comprehensive experiments are conducted in the CARLA simulation platform~\cite{dosovitskiy2017carla} to assess attack performance across various viewpoints, distances, weather conditions, and transferability across different object detectors.


\begin{table}[t]
\begin{center}
\begin{small}
\rowcolors{4}{gray!20}{white}
\resizebox{1\columnwidth}{!}{
\begin{tabular}{lrrrrrrrrr}
\toprule
\multirow{2}*{\textbf{Methods}} & \multicolumn{8}{c}{\textbf{Elevation angle}} & \multirow{2}*{$\bm{Avg}$}   \\ 
\cmidrule(lr){2-9}
                       & $\bm{0^\circ}$ &  $\bm{5^\circ}$  & $\bm{10^\circ}$      &$\bm{15^\circ}$    & $\bm{20^\circ}$      & $\bm{30^\circ}$        & $\bm{45^\circ}$           & $\bm{60^\circ}$    &                         \\ 
\midrule
Normal                   & 97.57    &99.23          & 99.56     & 98.23    &   99.78  & 96.24             & 85.42 & 45.30         & 90.17    \\ 
DAS                      & 93.37   &  95.58         & 99.45    & 93.37  &  98.34     & 95.58             & 73.48      & 45.30       & 86.81    \\ 
FCA                     & 93.37    &   73.48         & 64.09  & 65.75      &   70.17    & 39.78             & 10.50       & 3.31      & 52.56    \\ 
DTA                      & 76.80      & 64.09    &    60.22    & 72.38  &  82.87    & 56.91&  22.10      & 0.55     & 54.49    \\ 
ACTIVE                & 65.19    & 66.30      &    36.46  & 32.60    &  37.57   & 19.34          & 0.0      & 0.0       &    32.18   \\ 
RAUCA            & 38.67  &  23.20&      4.97 & 9.94  & 13.26    & 13.81       & 1.66    & 0       &  13.19   \\ 
\midrule
\textbf{Ours}   &  \textbf{12.53}   & \textbf{2.51} & \textbf{2.24} & \textbf{0.0} & \textbf{0} & \textbf{0} & \textbf{0} & \textbf{0}   & \textbf{2.16} \\
\bottomrule
\end{tabular}
}
\caption{Evaluation results in various elevation angles. Values are AP@0.5 (\%) of the target vehicle averaged across different azimuth angles with YOLOv3.}\label{tbl: perspectives}
\end{small}
\end{center}
\end{table}

\begin{figure}[t]
\vspace{-0.3cm}
\begin{center}
\centerline{\includegraphics[width=\columnwidth]{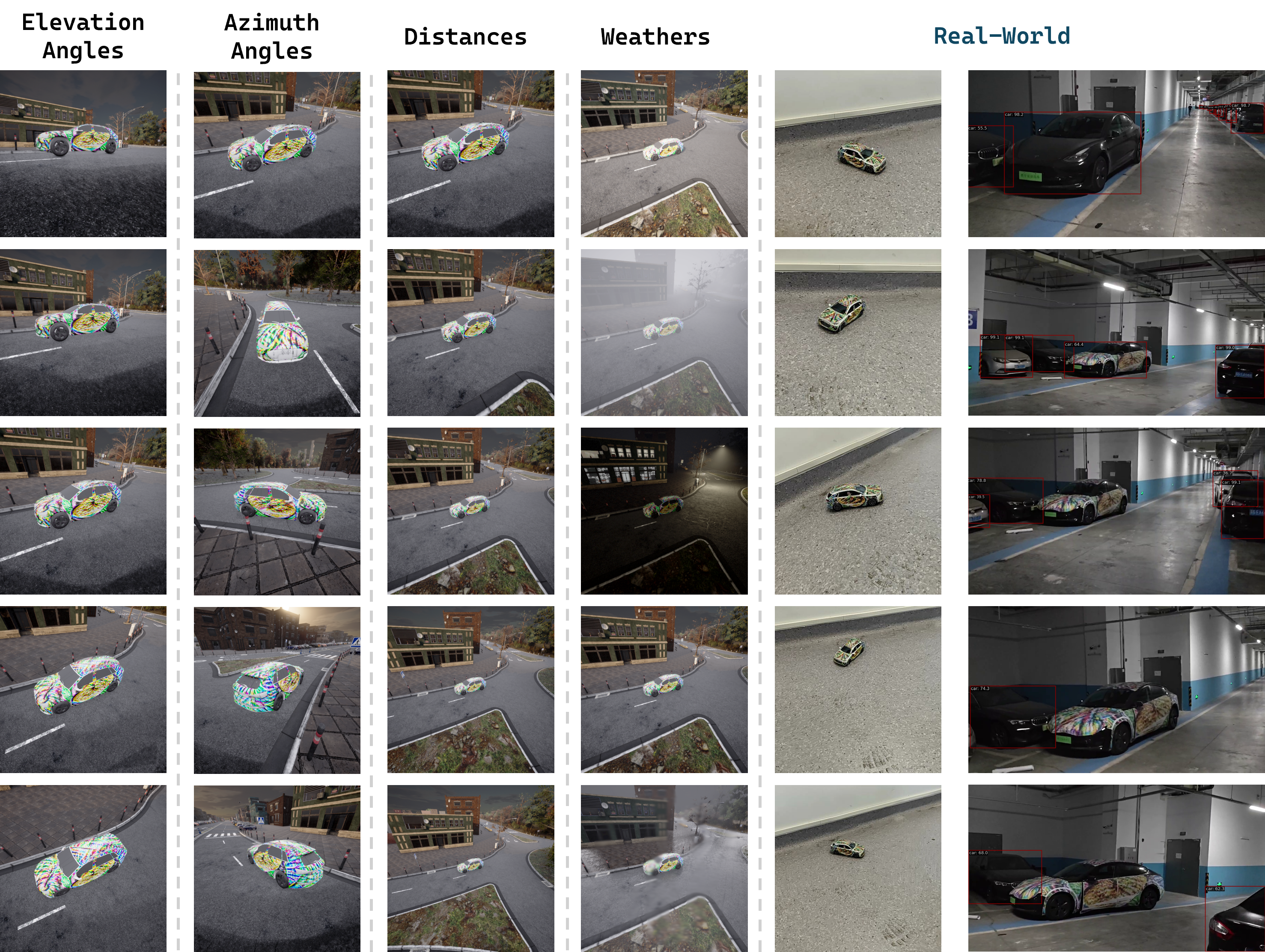}}
\caption{Samples under different evaluation conditions}
\vspace{-0.7cm}
\label{fig: eval_cond}
\end{center}
\end{figure}

\textbf{Robustness Across Multiple Angles}:  
We evaluated the effectiveness of adversarial attacks across $\bm{360^\circ}$ azimuth angles and elevation angles ranging from $\bm{0^\circ}$ to $\bm{60^\circ}$. For each elevation angle, we captured images encircling the target vehicle. The evaluation was conducted in a white-box scenario using the YOLOv3 detector, and the results are summarized in Table~\ref{tbl: perspectives}. The findings reveal that attacks are more challenging at lower elevation angles, as demonstrated by higher AP@50 scores across various methods. This difficulty stems from the presence of non-camouflaged features, such as vehicle tires, which act as strong cues for detection. In contrast, at higher elevation angles, where such robust features are often absent and only the camouflaged areas are visible, the detector’s performance weakens, leading to a higher attack success rate. Among all methods, our approach achieves the best results, consistently yielding the lowest AP@50 scores across all elevation angles. This showcases the effectiveness of our method in achieving a better overall balance across different perspectives. Specifically, compared to the state-of-the-art RAUCA~\cite{zhou2024rauca} approach, our method enhances attack performance by an average of $11.03\%$.

\textbf{Robustness Across Multiple Distances}:  
We conducted extensive experiments to assess performance across distances from 5 to 15 meters. Using the same protocol as for elevation evaluation, we averaged attack performance over full $\bm{360^\circ}$ azimuth and $\bm{0^\circ}$–$\bm{60^\circ}$ elevation angles. Results in Table~\ref{tbl: distances} reveal two key insights. First, although detection capability weakens with distance, attack performance doesn't necessarily improve due to loss of texture details at greater ranges, which can diminish camouflage effectiveness. Second, our method consistently achieves high attack success rates across all distances, outperforming existing approaches by 13.46\%, highlighting the robustness and adaptability of the proposed NGC strategy.


\begin{table}[t]

\begin{center}
\begin{small}
\rowcolors{4}{gray!20}{white}

\resizebox{1\columnwidth}{!}{
\begin{tabular}{lrrrrrr}
\toprule
\multirow{2}*{\textbf{Methods}} & \multicolumn{5}{c}{\textbf{Distance ($\bm{m}$)}} & \multirow{2}*{$\bm{Avg}$} \\ 
\cmidrule(lr){2-6}
& $\bm{5}$ & $\bm{7.5}$ &  $\bm{10}$          & $\bm{12.5}$      & $\bm{15}$       &                           \\ 
\midrule
Normal  &  60.36 & 90.54 &  90.17 & 86.33 &  84.25 & 82.33  \\ 
DAS~\cite{wang2021dual} &   63.61  &  89.44&  86.81 & 81.01 & 78.66 & 79.91 \\ 
FCA~\cite{wang2022fca}  &  51.59 & 68.65 &  52.56 & 32.94 &  44.82 & 50.11 \\ 
DTA~\cite{Suryanto_2022_CVPR}  &  35.22 &  68.03 &  54.49 & 33.77 & 44.41 & 47.18 \\ 
ACTIVE~\cite{Suryanto_2023_ICCV}  &  14.64  & 33.63& 32.18 & 25.34 & 37.02 & 28.56 \\ 
RAUCA~\cite{zhou2024rauca}  & 17.06 &  24.17 & 13.19 & 10.43 & 15.12 & 15.99 \\ 
\midrule
\textbf{Ours}  &   \textbf{1.24}  &   \textbf{2.48}  & \textbf{2.16}  &  \textbf{2.90}  &  \textbf{3.87}  &  \textbf{2.53} \\
\bottomrule
\end{tabular}
}
\caption{Evaluation results in various distances. Values are AP@0.5 (\%) of the target vehicle averaged across different elevation and azimuth angles with YOLOv3.}\label{tbl: distances}
\end{small}
\end{center}
\end{table}







    

\begin{table}[t]
\vspace{-0.3cm}
\begin{center}
\begin{small}
\rowcolors{4}{gray!20}{white}

\resizebox{\columnwidth}{!}{
\begin{tabular}{lccccccc}
\toprule
\multirow{2}*{\textbf{Methods}} & \multicolumn{5}{c}{\textbf{Weather Setting}} & \multirow{2}*{$\bm{Avg}$}\\ 
\cmidrule(lr){2-6}
    & Noon            & Sunset            & Night    & Foggy & Rainy   \\ 
\midrule
Normal & 92.27 & 90.17 &  92.34 &  79.01& 93.99& 89.56 \\
DAS~\cite{wang2021dual} & 86.67& 86.81 & 88.74& 81.70& 88.54& 86.51 \\
FCA~\cite{wang2022fca} & 66.99& 52.56 & 73.55&  59.81 & 64.64&  63.51\\
DTA~\cite{Suryanto_2022_CVPR}  & 69.48&  54.49 &  68.99&  55.32 & 67.40&  63.14\\
ACTIVE~\cite{Suryanto_2023_ICCV} & 50.28 & 32.18 &  38.40&  26.87& 43.30& 40.24\\
RAUCA~\cite{zhou2024rauca} & 29.63& 13.19 & 27.49 & 16.02& 24.86& 22.24 \\
\midrule
\textbf{Ours} & \textbf{5.25}& \textbf{2.16} & \textbf{1.86}&  \textbf{2.42}& \textbf{2.62} & \textbf{2.86} \\
\bottomrule
\end{tabular}
}
\caption{
Evaluation results under diverse weather conditions. Values are AP@0.5 (\%) of the target vehicle averaged across different azimuth angles with YOLOv3.}\label{tbl: illumination}
\vspace{-0.3cm}
\end{small}
\end{center}
\end{table}

\textbf{Robustness Under Different Weather Conditions}:  
Following previous studies~\cite{zhou2024rauca}, we evaluated our method under varying weather conditions, including \textit{Noon} (high illumination), \textit{Night} (low illumination), \textit{Sunset} (moderate illumination), as well as \textit{Foggy} and \textit{Rainy} conditions, which reduce visibility. The results, summarized in Table~\ref{tbl: illumination}, show that most existing methods are sensitive to these conditions. Extreme illumination (\textit{Noon} and \textit{Night}) leads to performance degradation due to overexposure or reduced visibility. Moreover, adverse weather conditions like \textit{Foggy} and \textit{Rainy} also impact performance by obscuring critical visual adversarial features. In contrast, our method demonstrates significantly improved robustness across all tested conditions. Notably, compared to state-of-the-art (SOTA) method RAUCA~\cite{zhou2024rauca}, our method achieves a consistent improvement, with an average gain of 19.38\% across all weather scenarios. Despite this, high illumination (\textit{Noon}) still poses challenges due to overexposure, providing an avenue for further enhancement.

\textbf{Transferability Across Different Object Detectors}:  
To evaluate the ability of our method to transfer across various object detectors, we carried out comprehensive experiments on detectors with diverse architectures, including single-stage, two-stage, and transformer-based models. These experiments compared our approach with several state-of-the-art (SOTA) methods. The outcomes, detailed in Table~\ref{tbl: transferability}, reveal notable variations in method performance across different detector types. Notably, our method not only surpasses existing SOTA approaches in the white-box scenario but also achieves consistently stronger results in most black-box tests, highlighting its superior transferability.



\begin{table}[t]
\begin{center}
\rowcolors{6}{gray!20}{white}

\resizebox{1\columnwidth}{!}{
\begin{tabular}{lccccccc}
\toprule
\multirow{2}*{\textbf{Methods}} & \multicolumn{2}{c}{\textbf{Single Stage}} &\multicolumn{2}{c}{\textbf{Two Stage}}  &\multicolumn{2}{c}{\textbf{Transformer}} \\ 
\cmidrule(lr){2-3} \cmidrule(lr){4-5}  \cmidrule(lr){6-7}
    & YOLOv3            & YOLOX             & FrRCN    & MkRCN & DETR & PVT   \\ 
\midrule
Normal & 90.17 & 95.30& 95.68 & 97.71& 97.56& 90.36 \\
DAS~\cite{wang2021dual} & 86.81& 88.60 & 85.84&87.50& 83.91& 79.63 \\
FCA~\cite{wang2022fca} & 52.56& 73.41& 67.95 &  71.34& 57.04&  54.97\\
DTA~\cite{Suryanto_2022_CVPR}  & 54.49&  79.63 &  66.16&  79.77 & 25.90&  65.19\\
ACTIVE~\cite{Suryanto_2023_ICCV} & 32.18 & 49.45 &  46.13&  50.48& 31.49& 41.09\\
RAUCA~\cite{zhou2024rauca} & 13.19& 30.80 & 23.34 & 33.77& 13.53& 35.43 \\
\midrule
\textbf{Ours} & \textbf{2.16}& \textbf{22.65} & \textbf{14.57}&  \textbf{28.11}& \textbf{6.28} & \textbf{28.25} \\
\bottomrule
\end{tabular}
}
\caption{Evaluating of transferability across various object detectors.}\label{tbl: transferability}
\end{center}
\end{table}

\begin{figure}[t]
\vspace{-0.3cm}
\begin{center}
\small
\begin{tabular}{ccc}
\includegraphics[width = 0.48\linewidth]{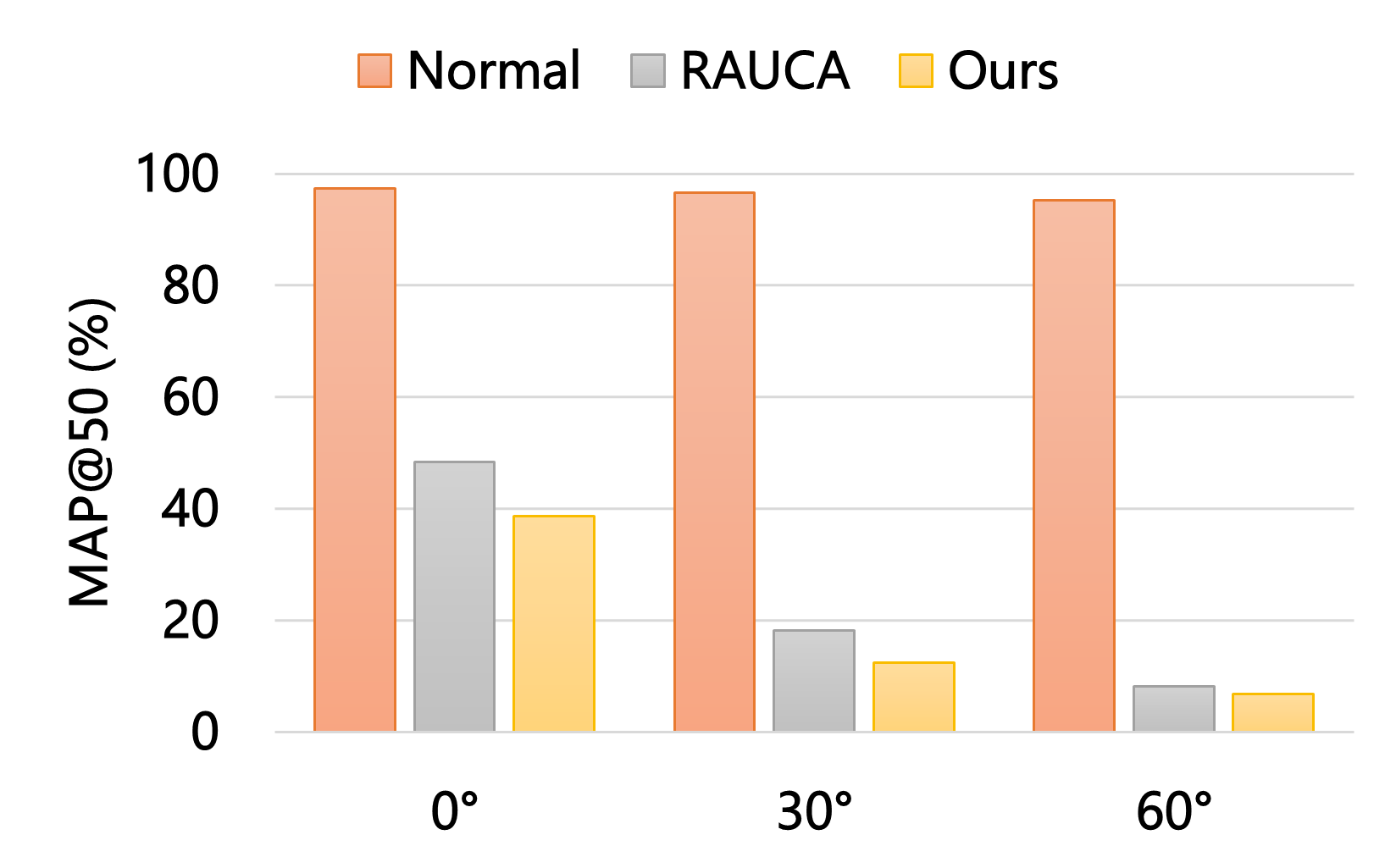}  &\hspace{-4.5mm}
\includegraphics[width = 0.48\linewidth]{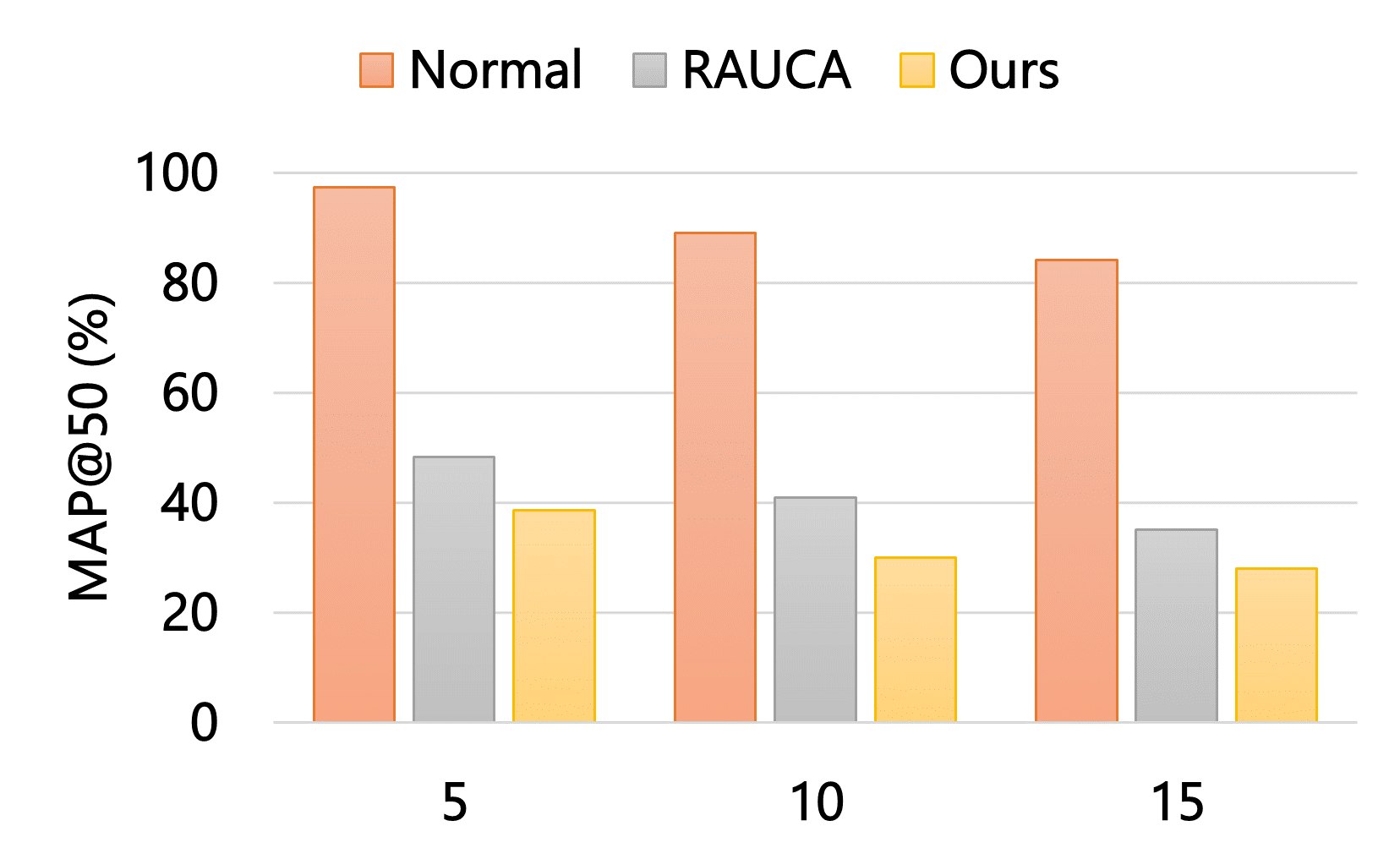}
\\
(a) Elevation Angle  & (b) Distance (Relative)
\end{tabular}
\end{center}
\caption{Evaluation results in a real-world setting, assessed under varying elevation angles (a) and distances (b).}
\vspace{-0.3cm}
\label{fig:real-world}
\end{figure}

\subsection{Evaluation in Real-World Settings}
In this section, we evaluate our method in real-world settings by printing and applying the optimized texture onto a car, as shown in Figure~\ref{fig: eval_cond}. The evaluation considers varying perspectives and distances. For perspectives, we tested three elevation angles and averaged results over a full $360^\circ$ azimuth. For distances, we assessed three distinct ranges. Results in Figure~\ref{fig:real-world} show that real-world conditions are more challenging than digital simulations. Lower elevation angles yield weaker performance, aligning with simulation trends. As distance increases, AP drops, indicating improved attack effectiveness—likely due to degraded detector performance at longer ranges. Additionally, while camouflage detail fades with distance, even robust features like uncovered tires become less detectable.

    

\begin{table}[b]
\vspace{-0.3cm}
\hspace{-0cm}
\centering
\small
\resizebox{0.7\columnwidth}{!}{%
\begin{tabular}{@{}llll@{}}
\toprule
 & FP+BP  & NGC  & LPGD  \\ \midrule
Complexity & / & $O(M\log N)$ & $O(m^2k)$  \\
Time cost (s) & 0.1163  &  +0.0209 & +0.0054   \\ \bottomrule
\end{tabular}%
}
\caption{Computational complexity and time cost.}
\label{tbl:computational complexity}
\vspace{-0.3cm}
\end{table} 

\subsection{Computational Complexity}
The NGC module has a complexity of $O(M \log N)$, where $M = |T_F|$ and $N = |T_U|$ denote the numbers of unsampled and sampled points, respectively. This stems from KD-Tree-based nearest neighbor search. The LPGD component introduces a complexity of $O(m^2k)$, due to gradient orthogonalization over a texture of size $m^2$ across $k$ optimization steps.
In runtime tests, NGC introduces an 18.0\% overhead (+0.0209s) per iteration, while LPGD adds 4.6\% (+0.0054s), compared to the base forward and backward time (0.1163s). 

\subsection{Ablation Studies}
In this section, we perform ablation studies to analyze the impact of our proposed NGC and LPGD strategies. As shown in Table~\ref{tbl:ablation_overall}, both components individually improve attack performance, with their combination yielding the best results.
\begin{table}[t]
\vspace{-0.0cm}
\hspace{-0cm}
\centering
\renewcommand{\arraystretch}{0.0}
\small
\resizebox{1\columnwidth}{!}{%
\begin{tabular}{@{}l|r|r|r|r@{}}
\toprule
Model &	Baseline &	w/ NGC & w/ LPGD & w/	NGC + LPGD \\
\midrule
Yolov3	& 11.75	& 3.45	& 8.63	& 2.16 \\
\midrule
Faster RCNN	& 40.95	& 18.33	& 31.63	& 14.57\\
\bottomrule
\end{tabular}%
}
\caption{Ablation study on our proposed components.}
\label{tbl:ablation_overall}
\vspace{-0.3cm}

\end{table} 
For NGC, we further examine the effect of the search radius $\tau$, which determines the furthest neighbor allowed for gradient propagation. For LPGD, we explore how the batch size used for gradient orthogonalization influences performance. To isolate the effects, we set the other component's effectiveness to zero during each study.

\begin{figure}[t]
\begin{center}
\small
\begin{tabular}{ccc}
\includegraphics[width = 0.48\linewidth]{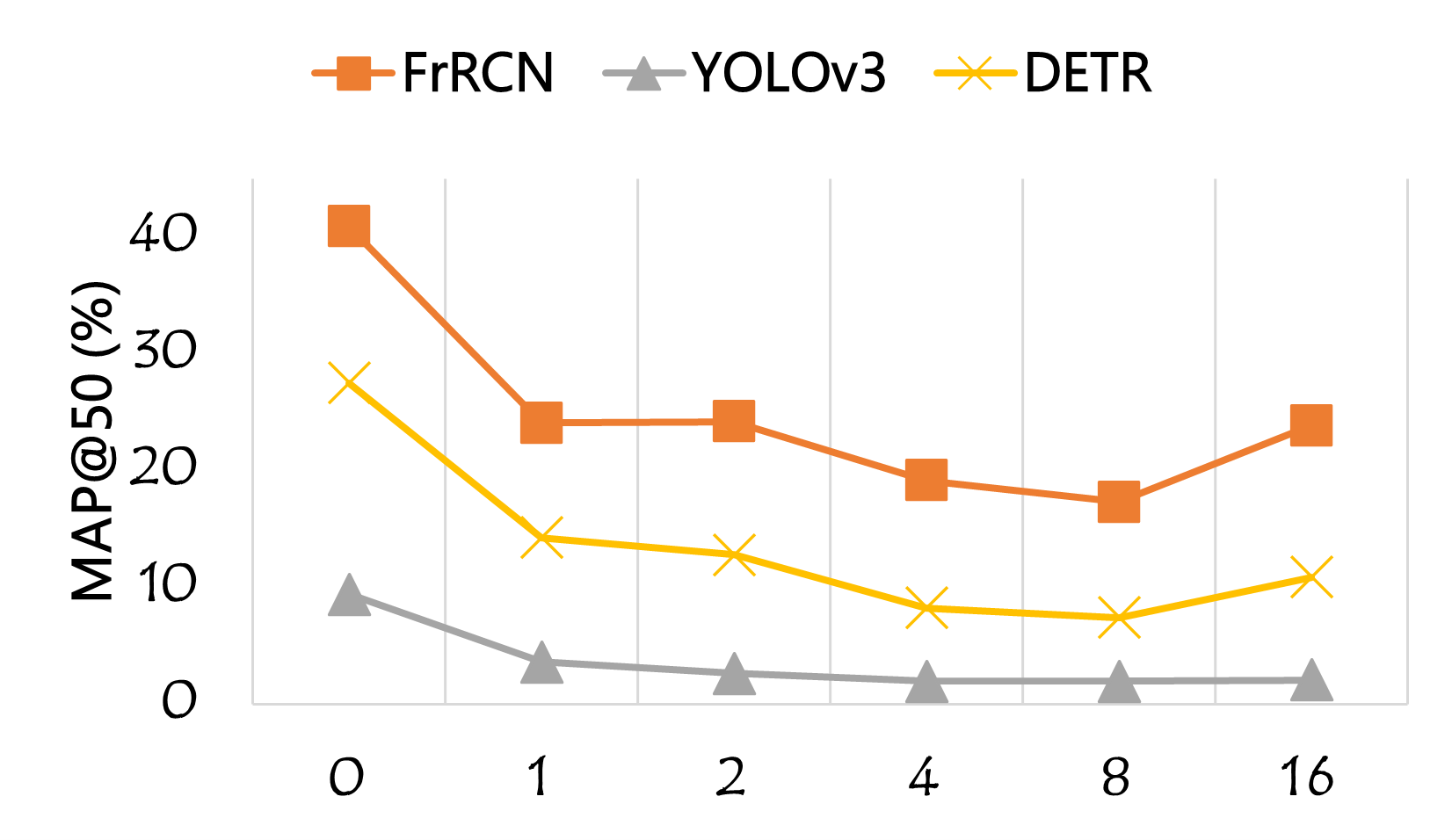}  &\hspace{-4.5mm}
\includegraphics[width = 0.48\linewidth]{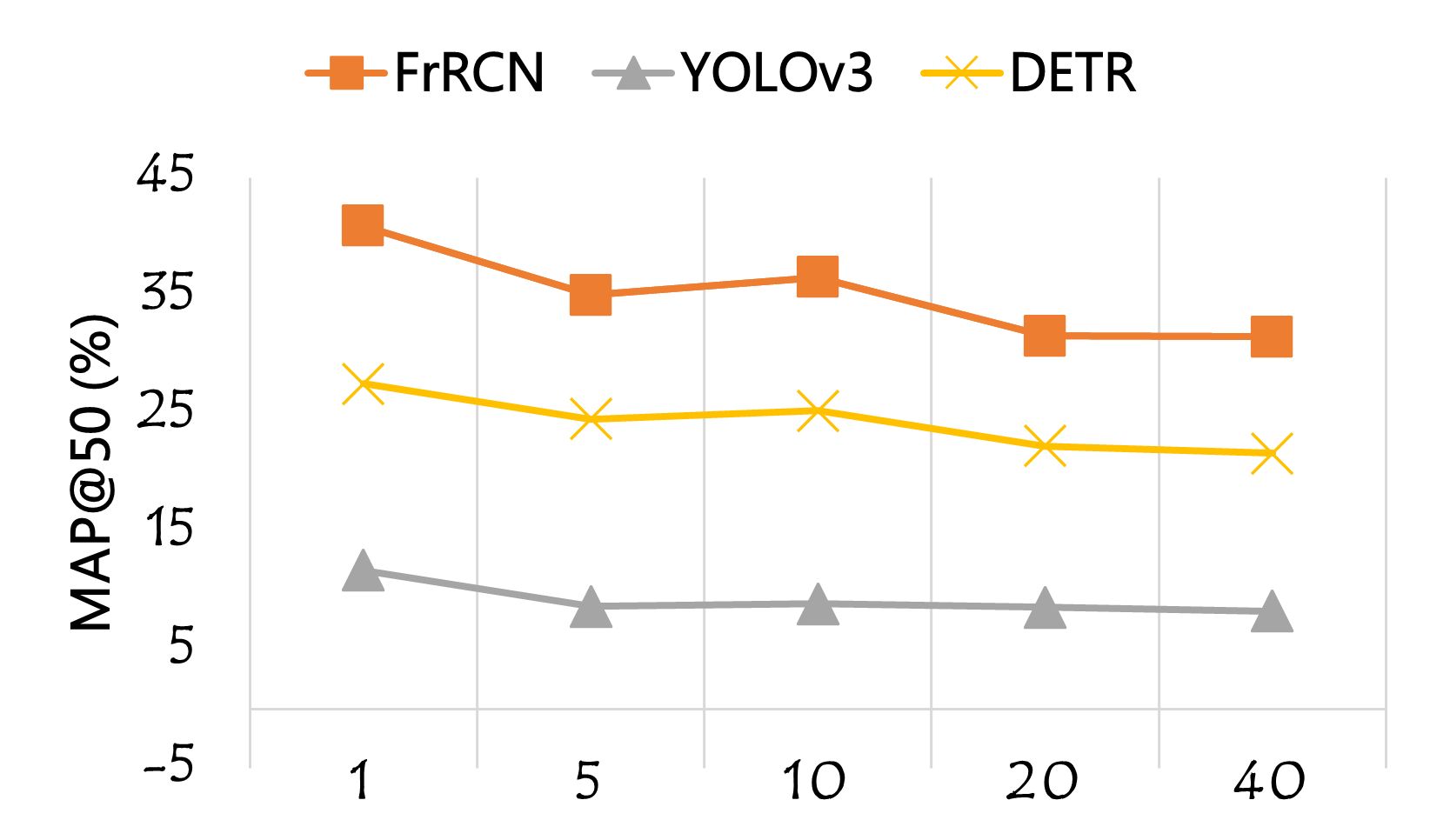}
\\
(a) Effect of search radius $\tau$ &\hspace{-4.5mm} (b) Effect of batchsize $k$ 
\end{tabular}
\end{center}
\vspace{-0.3cm}
\caption{Ablation study on (a) search radius in NGC and (b) batch size for gradients orthogonalization in LPGD.}
\label{fig:ablation}
\end{figure}

\textbf{Effect of Search Radius $\tau$}: We evaluate our NGC strategy using a range of search radius values $\tau$, from $0$ to $16$. The results, shown in Figure \ref{fig:ablation} (a), indicate that the attack performance initially improves as $\tau$ increases, reaching an optimal point, and then declines. This trend demonstrates the effectiveness of our method compared to the case where $\tau=0$. However, if $\tau$ becomes too large, it can cause gradients to propagate to areas beyond the immediate surface, introducing unnecessary noise to textures outside the target region. Consequently, selecting an appropriate $\tau$ is crucial to balancing local continuity and noise avoidance.

\textbf{Effect of Batch Size $k$}: We assess the LPGD strategy by varying the batch size from $1$ to $40$. The results, depicted in Figure \ref{fig:ablation} (b), show a trend of increasing attack performance with larger $k$. This validates our method's effectiveness in reducing conflicts by considering a broader global context and decorrelating redundancy and conflict between gradients.

\section{Conclusion}\label{sec: conclusion}
In this study, we identify two key challenges in physical adversarial camouflage: inconsistency in gradient sparsity and conflicting gradient updates. We introduce a novel framework that incorporates Nearest Gradient Calibration (NGC) and Loss-Prioritized Gradient Decorrelation (LPGD). NGC promotes gradient propagation from sampled to nearby unsampled texture points, ensuring local continuity across varying distances. LPGD prioritizes and orthogonalizes gradients to resolve redundancy and conflicts for gradients derived from different viewpoints. Our approach significantly enhances attack performance, robustness across diverse environments, and transferability across different detectors.
\newpage
\section*{Acknowledgements}
This work was supported by the Fundamental Research Funds for the Central Universities, Sun Yat-sen University under Grants No. 23xkjc010; the Shenzhen Science and Technology Program (No.KQTD20221101093559018).
\bibliographystyle{named}
\bibliography{ijcai25}

\end{document}